\documentclass[sigconf, screen]{acmart}

\usepackage{hyperref}
\usepackage{cleveref}
\usepackage{subcaption}
\usepackage[inline]{enumitem}

\AtBeginDocument{%
  }

\setcopyright{acmlicensed}
\copyrightyear{2025}
\acmYear{2025}
\acmDOI{XXXXXXX.XXXXXXX}

\acmConference[CIKM '25]{The 34th ACM International Conference on Information and Knowledge Management}{November 10--14, 2025}{Seoul, Korea}

\acmISBN{978-1-4503-XXXX-X/2018/06}

\acmSubmissionID{1104}

\begin{document}

\title[\texttt{ReDSM5}: A Reddit Dataset for DSM-5 Depression Detection]{\texttt{ReDSM5}: A Reddit Dataset for DSM-5 Depression Detection}

\author{Eliseo Bao}
\email{eliseo.bao@udc.es}
\orcid{0009-0000-8457-1115}
\affiliation{
  \institution{IRLab, CITIC, Universidade da Coruña}
  \city{A Coruña}
  \country{Spain}
}

\author{Anxo Pérez}
\email{anxo.pvila@udc.es}
\orcid{0000-0002-0480-006X}
\affiliation{
  \institution{IRLab, CITIC, Universidade da Coruña}
  \city{A Coruña}
  \country{Spain}
}

\author{Javier Parapar}
\email{javier.parapar@udc.es}
\orcid{0000-0002-5997-8252}
\affiliation{
  \institution{IRLab, CITIC, Universidade da Coruña}
  \city{A Coruña}
  \country{Spain}
}

\renewcommand{\shortauthors}{Bao et al.}

\begin{abstract}
Depression is a pervasive mental health condition that affects hundreds of millions of individuals worldwide, yet many cases remain undiagnosed due to barriers in traditional clinical access and pervasive stigma. Social media platforms, and Reddit in particular, offer rich, user-generated narratives that can reveal early signs of depressive symptomatology. However, existing computational approaches often label entire posts simply as \textit{depressed} or \textit{not depressed}, without linking language to specific criteria from the DSM-$5$, the standard clinical framework for diagnosing depression. This limits both clinical relevance and interpretability. To address this gap, we introduce \texttt{ReDSM5}, a novel Reddit corpus comprising $1484$ long-form posts, each exhaustively annotated at the sentence level by a licensed psychologist for the nine DSM-$5$ depression symptoms. For each label, the annotator also provides a concise clinical rationale grounded in DSM-$5$ methodology. We conduct an exploratory analysis of the collection, examining lexical, syntactic, and emotional patterns that characterize symptom expression in social media narratives. Compared to prior resources, \texttt{ReDSM5} uniquely combines symptom-specific supervision with expert explanations, facilitating the development of models that not only detect depression but also generate human-interpretable reasoning. We establish baseline benchmarks for both multi-label symptom classification and explanation generation, providing reference results for future research on detection and interpretability.
\end{abstract}

\ccsdesc[500]{Applied computing~Health informatics}
\ccsdesc[500]{Computing methodologies~Language resources}
\ccsdesc[300]{Information systems~Web mining}

\keywords{Depression symptom detection; Mental health; DSM-5; Social media; Health informatics; NLP; Language resources}


\maketitle

\section{Introduction and Motivation}
Depression is a leading cause of disability worldwide. The World Health Organization (WHO) reports that more than $280$ million people currently experience the disorder~\citep{healthorganization2023depressive}. Symptoms such as persistent sadness, anhedonia, and cognitive impairment reduce quality of life and, when left undetected, can lead to severe personal and societal costs~\citep{picardi2016randomised}. Although structured interviews and self-report inventories remain the diagnostic gold standard, many individuals do not seek help because of cost, limited availability, or stigma~\citep{gulliver2010perceived}. The popularity of social media creates an alternative source of behavioural evidence. Nearly three quarters of adults under $30$ actively use at least five platforms~\citep{researchcenter2024how}, and long-form communities like Reddit capture rich narratives of daily experience. Researchers have applied natural language processing (NLP) to these texts~\citep{bucur2025state}, identifying lexical, affective, and discourse markers that indicate depression risk~\citep{chandraguntuku2017detecting}. Despite encouraging accuracy, most existing systems are not grounded in recognised diagnostic criteria and rarely explain the reasoning behind their predictions, which limits clinical trust~\citep{uhauser2022promise}. 

A straightforward approach is to follow the same clinical guidelines that practitioners use during diagnostic interviews: the \textit{Diagnostic and Statistical Manual of Mental Disorders} (DSM-$5$), whose nine symptoms define a major depressive episode~\citep{psychiatricassociation2013diagnostic}. In routine clinical practice, each symptom is assessed individually, with clinicians probing for concrete verbal evidence before issuing a diagnosis. Reflecting that workflow, sentence-level annotation for the nine DSM-$5$ symptoms would allow models to learn \textit{what} specific clinical signals appear in a post and \textit{where} they occur. Equally important is \textit{why}: concise expert rationales expose the reasoning chain that links language to diagnostic criteria. Recent advances in Large Language Models (LLMs) make this granularity especially relevant, since chain-of-thought prompting shows that LLMs benefit from explicit intermediate reasoning when given appropriate supervision~\citep{wei2022chain-of-thought}. Fully leveraging this capacity in mental health NLP requires resources that pair fine-grained DSM-$5$ symptom labels with expert explanations.

To address this gap, we present \texttt{ReDSM5}, a dataset of $1484$ Reddit\footnote{\url{https://www.reddit.com/}} posts annotated at the sentence level for DSM-$5$ symptoms. A licensed psychologist annotated every sentence for the presence or absence of the nine DSM-$5$ depression symptoms and wrote a brief clinical rationale for each label. The dataset\footnote{\url{https://huggingface.co/datasets/irlab-udc/redsm5}} and supporting code\footnote{\url{https://github.com/eliseobao/redsm5}} are released under a research license to encourage reproducibility and further research in explainable depression detection.

The remainder of this paper is organized as follows: \Cref{sec:data_acquisition_and_annotation} details the data acquisition phase and the annotation of the \texttt{ReDSM5} dataset. \Cref{sec:dataset_overview} presents the exploratory analysis conducted on this new resource, including both lexical and psycholinguistic approaches. In \Cref{sec:baselines}, we describe our baseline setup for depression symptom detection and explanation generation using the \texttt{ReDSM5} dataset. Finally, \Cref{sec:discussion_and_conclusions} concludes the paper by highlighting limitations and outlining directions for future work.

\section{Data Acquisition and Annotation}
\label{sec:data_acquisition_and_annotation}

We started from \textit{DepreSym}~\citep{prez2025depresym}, a collection of $21580$ Reddit sentences labeled for the $21$ symptoms of the \textit{Beck Depression Inventory} (BDI-II), a widely used self-report questionnaire for assessing depression severity~\citep{beck1996beck}. The \textit{DepreSym} resource had already been manually annotated by a team of three clinical experts as part of its preparation for the eRisk workshops\footnote{\url{https://erisk.irlab.org/}}. Since \textit{DepreSym} is one of our previous projects, we had direct access to its underlying data. This allowed us to reconstruct complete Reddit posts from the sentence-level files. After de-duplication, the process yielded $1484$ unique posts that became the raw pool for \texttt{ReDSM5}. Regarding the annotation process, the original \textit{DepreSym} labels were preserved only as non-binding cues. A custom web interface displayed each reconstructed post in full, highlighted the validated BDI-II tags, and provided the DSM-$5$ checklist~\citep{psychiatricassociation2013diagnostic}. Building on the existing expert annotations, a fourth licensed psychologist reviewed every post, mapped relevant BDI-II cues to their DSM-$5$ counterparts, discarded noisy labels, and marked sentences that evidenced or contradicted any of the nine DSM-$5$ depression symptoms. For each decision, the annotator added a concise clinical rationale. The outcome is a sentence-level resource whose labels and explanations align directly with DSM-$5$ diagnostic practice.

\section{Dataset Overview}
\label{sec:dataset_overview}

We begin by reporting core corpus statistics, including post counts, symptom coverage, and length distributions. We then analyze the dataset along several linguistic dimensions, including lexical and syntactic patterns, sentiment and emotion profiles, named-entity usage, topic structure, and grammar.

\paragraph{\textbf{Main Statistics}}
\label{paragraph:general_statistics}

\Cref{tab:redsm5_statistics} reports counts and length measures. The dataset comprises a total of $1484$ posts. On average, each post exhibits $1.39$ symptoms, reflecting that many posts discuss only a single symptom, while a smaller subset describe multiple expressions of depressive symptoms. We also identify $392$ \textit{negatives}, posts that were initially flagged during the BDI-II annotation process but were judged by our expert annotator to contain no DSM-$5$ depression symptoms. These cases reflect content that may be relevant under the broader scope of BDI-II criteria but do not meet the stricter symptom definitions specified by DSM-$5$. Post lengths vary considerably, from as few as $2$ words up to $6990$ words, with a mean of $294.7$ words per post, indicating both succinct expressions of distress and extended personal narratives.

\begin{table}
  \centering
  \caption{Main statistics of the \texttt{ReDSM5} collection.}
  \label{tab:redsm5_statistics}
  \begin{tabular}{@{}l r l r@{}}
    \toprule
    Statistic & Value & Statistic & Value \\
    \midrule
    \# Posts                        & $1484$      & Avg.  symptoms/post   & $1.39$   \\
    \# \textit{Negatives}           & $392$       & Avg.  words post      & $294.77$ \\
    Min. words post                 & $2$         & Max. words post       & $6990$   \\
    \bottomrule
  \end{tabular}
\end{table}

Next, \Cref{tab:redsm5_distribution_of_symptoms} shows the absolute frequency of each DSM-$5$ symptom across all posts.\textit{ Depressed mood} is the most commonly annotated symptom ($328$ occurrences), closely followed by \textit{worthlessness} ($311$), while \textit{psychomotor alteration} appears least frequently ($35$ occurrences). \textit{Suicidal thoughts}, a critical red‐flag indicator, occur in $165$ posts, showing the dataset's clinical relevance. Overall, the label distribution shows a long‐tail profile: a handful of symptoms dominate the annotation space, whereas others are relatively rare.

\begin{table}
  \centering
  \caption{Distribution of symptoms of depression and their occurrence in the \texttt{ReDSM5} collection.}
  \label{tab:redsm5_distribution_of_symptoms}
  \begin{tabular}{@{}l r l r@{}}
    \toprule
    Symptom & Occ. & Symptom & Occ. \\
    \midrule
    Depressed mood          & $328$ & Fatigue                & $124$ \\
    Anhedonia               & $124$ & Worthlessness          & $311$ \\
    Changes in appetite     &  $44$ & Cognitive issues       &  $59$ \\
    Sleep issues            & $102$ & Suicidal thoughts      & $165$ \\
    Psychomotor alteration  &  $35$ & \multicolumn{2}{c}{\;} \\ 
    \bottomrule
  \end{tabular}
\end{table}

\paragraph{\textbf{Named Entity Recognition (NER)}}
\label{paragraph:ner}

To better understand how depressive symptoms are linguistically framed, we applied NER to extract structured references such as people, time expressions, quantities, and locations. This allowed us to examine what individuals focus on when describing their experiences and how different symptoms are grounded in specific contexts. Temporal references deserve special attention, since detecting explicit or implicit time spans is a prerequisite to justify a clinical diagnosis~\citep{beck1996beck,cooper2018diagnosing}.

\begin{figure*}[ht]
  \centering
  \includegraphics[width=1.0\linewidth]{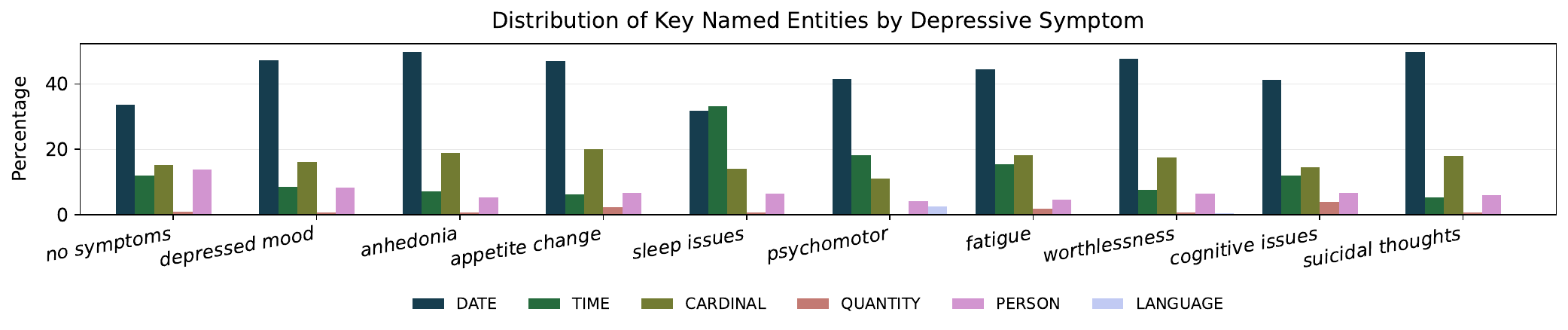}
  \caption{Distribution of selected named entity types across depressive symptoms. For clarity, the figure displays only a subset of the most informative entity categories.}
  \Description{Bar chart comparing the percentage of selected named entity types by symptom category.} 
   \label{fig:ner}
\end{figure*}

As shown in \Cref{fig:ner}, temporal entities, particularly \texttt{DATE}, \texttt{TIME}, and \texttt{CARDINAL}, were dominant across several categories, most notably in \textit{depressed mood}, \textit{anhedonia}, and \textit{suicidal thoughts}. This suggests that users often describe these symptoms in relation to specific durations or recurring moments. Posts tagged with \textit{sleep issues} contained a high proportion of \texttt{TIME} references, consistent with detailed descriptions of disrupted sleep patterns. Mentions of \texttt{QUANTITY} and \texttt{CARDINAL} were more frequent in \textit{appetite change}, likely reflecting weight fluctuations or food intake. \textit{Psychomotor} symptoms showed an increase in \texttt{LANGUAGE} entities, possibly indicating changes in speech or cognitive processes.

\paragraph{\textbf{Topic Modeling}}
\label{paragraph:topic_modeling}

\begin{figure}[t]
  \centering
  \begin{subfigure}[t]{\linewidth}
    \centering
    \includegraphics[width=1.0\linewidth]{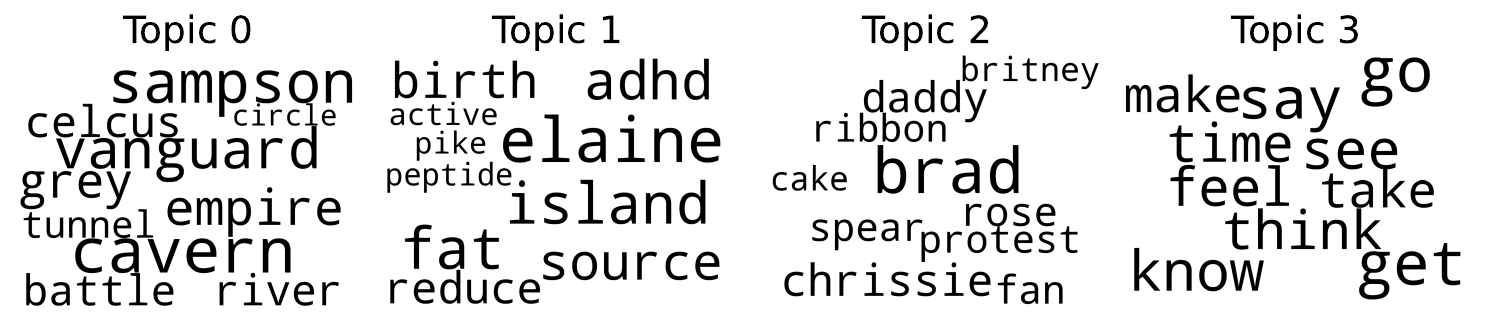}
    \label{fig:wordcloud_NO_symptoms}
  \end{subfigure}
  \begin{subfigure}[t]{\linewidth}
    \centering
    \includegraphics[width=1.0\linewidth]{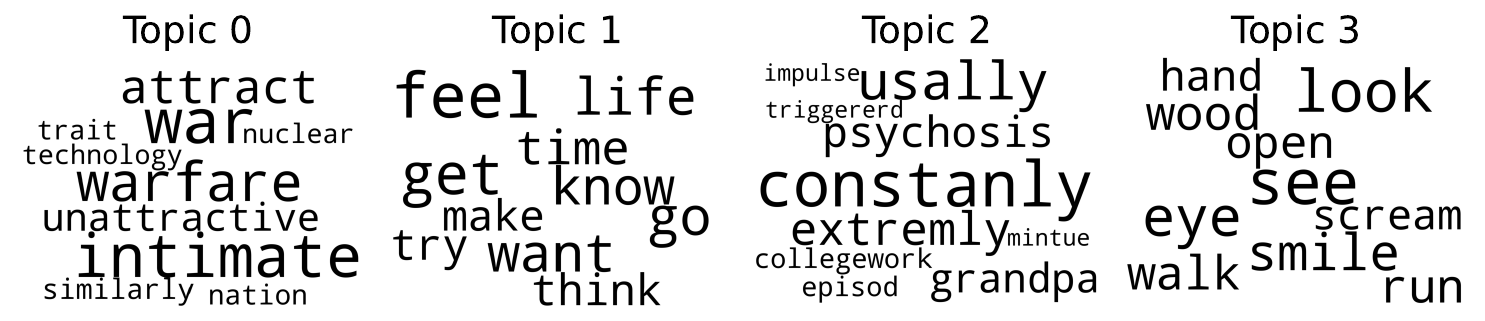}
    \label{fig:wordcloud_symptoms}
  \end{subfigure}
  \caption{Word clouds of the top terms for each LDA-inferred topic, shown separately for posts that were not tagged with any depression symptom (top) and those that were (bottom).}
  \label{fig:wordclouds}
  \Description{Two vertically stacked word clouds. Top: posts without depressive symptoms, featuring large words such as sampson, vanguard, cavern, empire, brad, and elaine, reflecting community and niche discourse. Bottom: posts with depressive symptoms, featuring prominent terms like feel, life, get, know, want, think; intensifiers such as usually, constantly, extremely, psychosis, triggered; and body-related nouns and action verbs such as hand, eye, walk, run, scream, indicating introspective and affective language.}
\end{figure}

To uncover latent thematic structures, we applied Latent Dirichlet Allocation (LDA)~\citep{blei2003latent} separately to the \textit{no-symptoms} (\textit{negatives}) and \textit{symptoms} subsets and then visualized the top terms for each topic with word clouds (\Cref{fig:wordclouds}). In the \textit{no-symptoms} clouds, the most salient tokens are proper nouns or niche vocabulary, for example \textit{sampson}, \textit{vanguard}, \textit{cavern}, \textit{empire}, \textit{brad}, and \textit{elaine}. These words point to gaming, fandom, or hobby discussions and show little emotional content. By contrast, the \textit{symptoms} clouds foreground introspective and affective language. Topic~1 centres on core predicates such as \textit{feel}, \textit{life}, \textit{get}, \textit{know}, \textit{want}, and \textit{think}. Topic~2 combines intensifiers and clinical terms such as \textit{usually}, \textit{constantly}, \textit{extremely}, \textit{psychosis}, and \textit{triggered}. Topic~3 highlights body-related nouns and action verbs: \textit{hand}, \textit{eye}, \textit{walk}, \textit{run}, and \textit{scream}. Taken together, these contrasts confirm that LDA separates general, community-oriented discussion from the self-referential, emotionally loaded language characteristic of depressive expression in our corpus.

\paragraph{\textbf{Psycholinguistics and Emotions}}
\label{paragraph:psycholinguistics_and_emotions}

We utilized the NRC Emotion Lexicon~\citep{mohammad2012crowdsourcing}, grounded in Plutchik's wheel of emotions~\cite{plutchik1980emotion}, to conduct an emotion-based analysis of the texts. Each post was evaluated for the presence of words associated with eight primary emotions and two general sentiments (\texttt{positive}, \texttt{negative}). This analysis provides insight into the emotional tone of the content and how it varies across individual symptoms in our multi-label setup.

\begin{figure}[ht]
  \centering
  \includegraphics[width=0.7\linewidth]{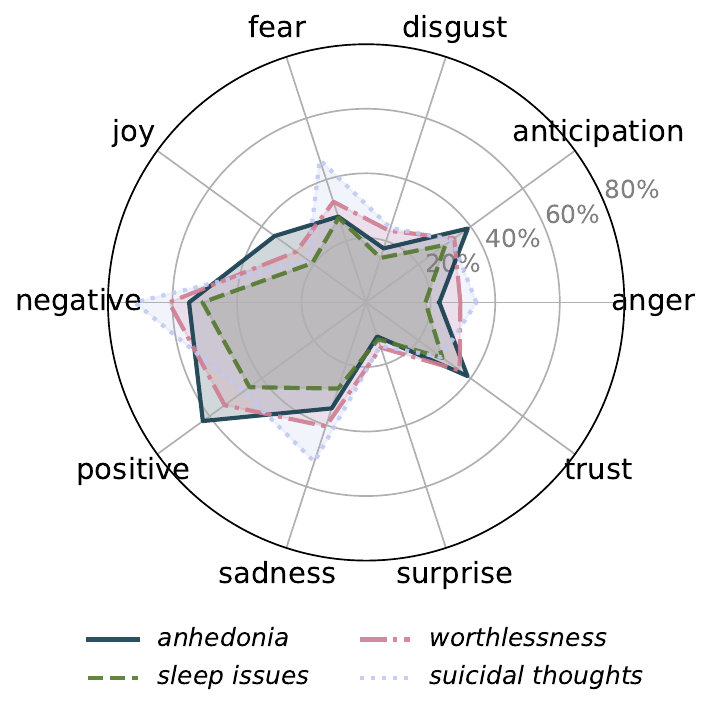}
  \caption{Emotion distribution across the most informative subset of symptoms. Each line represents the average proportion of emotion-related words for a given symptom class.}
  \label{fig:plutchik}
  \Description{Radar chart showing emotional categories across symptom classes.}
\end{figure}

As shown in \Cref{fig:plutchik}, posts labeled with \textit{suicidal thoughts} exhibited the highest proportions of \texttt{fear} and \texttt{sadness}, alongside the strongest overall \texttt{negative} sentiment. \textit{Anhedonia} posts registered the lowest levels of \texttt{joy} and \texttt{trust}, reflecting emotional withdrawal, but they showed a relative bump in \texttt{positive} terms compared to \textit{suicidal thoughts}, suggesting occasional neutral or hopeful language. \textit{Sleep issues} displayed a more muted profile: lower peaks in both \texttt{fear} and \texttt{sadness} and the smallest \texttt{negative} footprint overall. \textit{Worthlessness} fell between these extremes, with elevated \texttt{negative} and \texttt{sadness} but moderate \texttt{anger} and \texttt{trust}. Together, these patterns confirm that specific DSM-$5$ symptoms carry distinct emotional signatures in social-media discourse.

\paragraph{\textbf{Grammar and Pronoun Usage}}
\label{paragraph:grammar_and_prnoun_usage}

Finally, we examined the use of personal pronouns and verb tenses to identify grammatical patterns potentially associated with communicative intent. The full distribution is reported in \Cref{tab:grammar_usage}. Across all symptom categories, there was a strong predominance of \texttt{first person singular} (FPS) pronouns. This was especially notable in \textit{anhedonia} ($69.86$\%) and \textit{appetite change} ($68.98$\%), suggesting highly introspective discourse. In contrast, \texttt{first person plural} (FPP) and \texttt{third person plural} (TPP) pronouns appeared infrequently, indicating limited reference to shared or external experiences. Posts labeled with \textit{no symptoms} contained more \texttt{third person singular} (TPS) references ($30.37$\%) and slightly higher \texttt{second person} (SP) use, potentially reflecting more narrative or outward-focused content.

Verb tense usage also revealed important differences. Posts labeled with \textit{anhedonia}, \textit{cognitive issues}, and \textit{suicidal thoughts} showed high levels of \texttt{present tense} verbs (above $55$\%), indicating a focus on current or persistent mental states. In contrast, \textit{no symptoms} posts relied more heavily on the \texttt{past tense} ($49.99$\%), which may reflect more distanced or retrospective narration.

\begin{table}
  \footnotesize
  \centering
  \caption{Verb-tense and personal pronoun distribution across symptom categories (percentages).}
  \label{tab:grammar_usage}
  \begin{tabular}{@{}lccc ccccc@{}}
    \toprule
     & \multicolumn{3}{c}{Verb Tense (\%)} & \multicolumn{5}{c}{Pronouns (\%)}\\
    \cmidrule(lr){2-4}\cmidrule(l){5-9}
    Symptom & Pst. & Prs. & Fut. & FPS & FPP & SP & TPS & TPP\\
    \midrule
    No Sympt.        & $49.99$ & $38.96$ & $11.04$ & $51.84$ & $5.21$ & $8.41$ & $30.37$ & $4.16$ \\
    Dep.\ Mood       & $42.74$ & $46.40$ & $10.86$ & $63.52$ & $3.84$ & $6.11$ & $22.43$ & $4.09$ \\
    Anhedonia        & $26.51$ & $61.83$ & $11.67$ & $69.86$ & $1.69$ & $6.57$ & $19.18$ & $2.69$ \\
    App.\ Change     & $40.33$ & $50.23$ & $ 9.44$ & $68.98$ & $1.17$ & $5.38$ & $19.95$ & $4.52$ \\
    Sleep Issues     & $46.19$ & $43.03$ & $10.78$ & $65.23$ & $2.53$ & $7.29$ & $21.11$ & $3.84$ \\
    Psychomotor      & $40.55$ & $47.81$ & $11.64$ & $64.60$ & $2.48$ & $7.13$ & $22.23$ & $3.56$ \\
    Fatigue          & $39.37$ & $48.04$ & $12.59$ & $63.01$ & $4.36$ & $7.79$ & $21.25$ & $3.58$ \\
    Worthlessness    & $36.32$ & $52.13$ & $11.55$ & $67.50$ & $2.61$ & $6.09$ & $20.94$ & $2.85$ \\
    Cognitive        & $31.31$ & $55.98$ & $12.71$ & $68.30$ & $1.66$ & $3.32$ & $24.23$ & $2.49$ \\
    Suicidal         & $30.55$ & $57.44$ & $12.01$ & $66.84$ & $1.21$ & $6.21$ & $22.69$ & $3.05$ \\
    \bottomrule
  \end{tabular}
\end{table}

\section{Experiments}
\label{sec:baselines}

To provide solid reference points for future work with \texttt{ReDSM5}, we establish baselines in two complementary tasks:
\begin{enumerate*}[label=(\roman*)]
    \item \textit{multi-label symptom classification}, where models predict DSM-$5$ symptom labels;
    \item \textit{explanation generation}, where a LLM produces sentence-level rationales that we score independently of the classification decision.
\end{enumerate*} Both tasks share the same train/test split, with $80$\% of the data used for training and the remaining $20$\% reserved for evaluation. The released code contains runnable examples for every model and metric reported.

\paragraph{\textbf{Multi-label symptom classification}}
\label{paragraph:multi_label_symptom_classification}

We evaluate traditional models such as a linear Support Vector Machine (SVM) on TF-IDF unigrams. Additionally, inspired by recent studies~\citep{husseiniorabi2018deep,yang2023towards}, we include a ten-epoch Convolutional Neural Network (CNN), a five-epoch fine-tuned \texttt{bert-base-uncased}, and a four-epoch fine-tuned LLM, specifically \texttt{llama-3.2-1b}. Hyper-parameter details are included in the repository. The results, summarized in \Cref{tab:baselines_classification}, reveal large performance gaps among the models for the proposed multi-label classification. The SVM and CNN baselines yield modest accuracy scores of $0.22$ and $0.14$, respectively, with micro F1 scores of $0.39$ and $0.25$ and weighted F1 scores of $0.37$ and $0.23$. In contrast, fine-tuning BERT produces substantial gains, achieving an accuracy of $0.41$ along with micro, macro, and weighted F1 scores of $0.51$, $0.36$, and $0.48$. The fine-tuned LLM further improves these results, achieving an accuracy of $0.45$ and F1 scores of $0.54$ (micro), $0.49$ (macro), and $0.53$ (weighted). These scores suggest that large pre-trained transformers, especially under task-specific tuning, are far better at capturing the nuanced linguistic cues associated with depressive symptoms.

\begin{table}
  \small
  \caption{Baseline results for multi-label symptom classification on the \texttt{ReDSM5} dataset.}
  \label{tab:baselines_classification}
  \begin{tabular}{ccccc}
    \toprule
    Method & $\text{F1}_{micro}$ & $\text{F1}_{macro}$ & $\text{F1}_{wgt.}$ & Acc. \\
    \midrule
    SVM     & $0.39$            & $0.28$            & $0.37$            & $0.22$ \\
    CNN     & $0.25$            & $0.19$            & $0.23$            & $0.14$ \\
    BERT    & $0.51$            & $0.36$            & $0.48$            & $0.41$ \\
    LLM     & $\textbf{0.54}$   & $\textbf{0.49}$   & $\textbf{0.53}$   & $\textbf{0.45}$ \\
  \bottomrule
\end{tabular}
\Description{Table reporting the classification performance of four models on the ReDSM5 multi-label symptom detection task. SVM achieves micro-F1 of 0.39 and accuracy of 0.22. CNN performs worse with micro-F1 of 0.25 and accuracy of 0.14. Fine-tuned BERT reaches micro-F1 of 0.51 and accuracy of 0.41. The best results are obtained with fine-tuned LLM, achieving micro-F1 of 0.54 and accuracy of 0.45.}
\end{table}

\paragraph{\textbf{Explanation generation}}
\label{paragraph:explanation_generation}

We prompt \texttt{gemma-3 27B} in a few-shot setup, selecting examples from the training split so that the model produces DSM-$5$ aligned rationales in the same structured format used in the \texttt{ReDSM5} dataset. We then evaluate the output with metrics that prioritise meaning rather than surface form. Classical options such as BLEU~\citep{papineni2002bleu} or ROUGE~\citep{lin2004rouge}, although widespread in text generation, reward exact n-gram overlap and therefore penalise clinically valid paraphrases or reorderings, which makes them poorly suited for free-form explanations. We prefer two complementary, semantics-based approaches. First, we compute cosine similarity between \texttt{nomic-embed-text} embeddings of candidate and reference explanations, replicating the core step in BERTScore~\citep{zhang2020bertscore}. The average similarity is $0.78$, comfortably above the Sentence-BERT~\citep{reimers2019sentence-bert} threshold of $0.75$, which means that most generated explanations are near-paraphrases of the expert rationales. Second, we use \texttt{deepseek-r1 8B} as an automatic judge~\citep{zheng2023judging} that scores each explanation for clinical accuracy, coverage and clarity with weights of $0.40$, $0.30$ and $0.30$ respectively, producing an overall mean of $0.62$ on a $0$-$1$ scale\footnote{The full prompt templates and evaluation scripts are available in the repository.}. Together, these results show that even without any task specific fine-tuning a modern $27$B parameter LLM already produces explanations that are strongly aligned with expert annotations and clinically adequate in most cases.

\section{Discussion and Conclusions}
\label{sec:discussion_and_conclusions}

In this work, we introduced \texttt{ReDSM5}, a curated collection of Reddit posts annotated at the sentence level for the nine DSM-$5$ depression symptoms by a licensed psychologist. We argue that the alignment of automated systems with clinical diagnostic standards is a critical requirement for their acceptance in professional settings, and that datasets constructed following such criteria are a fundamental step toward achieving this goal. In addition to presenting the dataset, we conducted an exploratory analysis of its linguistic and emotional characteristics, providing further insight into how depressive symptoms are expressed in social media narratives. We also established baseline performance for both multi-label symptom classification and explanation generation, providing reference values for future research on detection and interpretability. Several limitations remain. Expanding the dataset would improve coverage, but this is a time-intensive process that, in our view, should continue to rely on professional annotation rather than semi-automated methods. Extending beyond English Reddit posts to other languages and platforms is also a necessary direction for future work.

\section*{Ethical Statement}
\label{sec:ethical_statement}

The data used to curate this resource was obtained from publicly accessible repositories, in compliance with the exempt status outlined in Title 45 CFR \S 46.104. All datasets were utilized in strict accordance with their respective data usage policies.

\begin{acks}  
This work was supported by the project PID2022-137061OB-C21 (MCIN/AEI/10.13039/501100011033, Ministerio de Ciencia e Innovación, ERDF, \textit{A way of making Europe} by the European Union); the Consellería de Educación, Universidade e Formación Profesional, Spain (grant number ED481A-2024-079 and accreditations 2019-2022 ED431G/01 and GRC ED431C 2025/49); and the European Regional Development Fund, which supports the CITIC Research Center.  
\end{acks}

\section*{GenAI Usage Disclosure}
\label{sec:genai_usage_disclosure}
During the preparation of this manuscript, generative AI tools were employed solely for light editing purposes, including proofreading, grammar correction, vocabulary improvement, and overall language polishing. All substantive ideas, analyses, experiments, and written content were created by the co-authors without direct text generation from any AI model.

\balance
\bibliographystyle{ACM-Reference-Format}
\bibliography{references}


\begin{thebibliography}{22}


\ifx \showCODEN    \undefined \def \showCODEN     #1{\unskip}     \fi
\ifx \showISBNx    \undefined \def \showISBNx     #1{\unskip}     \fi
\ifx \showISBNxiii \undefined \def \showISBNxiii  #1{\unskip}     \fi
\ifx \showISSN     \undefined \def \showISSN      #1{\unskip}     \fi
\ifx \showLCCN     \undefined \def \showLCCN      #1{\unskip}     \fi
\ifx \shownote     \undefined \def \shownote      #1{#1}          \fi
\ifx \showarticletitle \undefined \def \showarticletitle #1{#1}   \fi
\ifx \showURL      \undefined \def \showURL       {\relax}        \fi
\providecommand\bibfield[2]{#2}
\providecommand\bibinfo[2]{#2}
\providecommand\natexlab[1]{#1}
\providecommand\showeprint[2][]{arXiv:#2}

\bibitem[{American Psychiatric Association}(2013)]%
        {psychiatricassociation2013diagnostic}
\bibfield{author}{\bibinfo{person}{{American Psychiatric Association}}.} \bibinfo{year}{2013}\natexlab{}.
\newblock \bibinfo{booktitle}{\emph{Diagnostic and statistical manual of mental disorders: DSM-5{\texttrademark}} (\bibinfo{edition}{5th} ed.)}.
\newblock \bibinfo{publisher}{American Psychiatric Publishing, Inc.}, \bibinfo{address}{Arlington, VA, US}. 947 pages.
\newblock
\showISBNx{978-0-89042-554-1 (Hardcover); 978-0-89042-555-8 (Paperback)}
\href{https://doi.org/10.1176/appi.books.9780890425596}{doi:\nolinkurl{10.1176/appi.books.9780890425596}}


\bibitem[Beck et~al\mbox{.}(1996)]%
        {beck1996beck}
\bibfield{author}{\bibinfo{person}{Aaron~T. Beck}, \bibinfo{person}{R.~A. Steer}, {and} \bibinfo{person}{G. Brown}.} \bibinfo{year}{1996}\natexlab{}.
\newblock \bibinfo{title}{Beck Depression Inventory–II}.
\newblock
\href{https://doi.org/10.1037/t00742-000}{doi:\nolinkurl{10.1037/t00742-000}}


\bibitem[Blei et~al\mbox{.}(2003)]%
        {blei2003latent}
\bibfield{author}{\bibinfo{person}{David~M. Blei}, \bibinfo{person}{Andrew~Y. Ng}, {and} \bibinfo{person}{Michael~I. Jordan}.} \bibinfo{year}{2003}\natexlab{}.
\newblock \showarticletitle{Latent dirichlet allocation}.
\newblock \bibinfo{journal}{\emph{J. Mach. Learn. Res.}} \bibinfo{volume}{3}, \bibinfo{number}{null} (\bibinfo{date}{March} \bibinfo{year}{2003}), \bibinfo{pages}{993–1022}.
\newblock
\showISSN{1532-4435}


\bibitem[Bucur et~al\mbox{.}(2025)]%
        {bucur2025state}
\bibfield{author}{\bibinfo{person}{Ana-Maria Bucur}, \bibinfo{person}{Andreea-Codrina Moldovan}, \bibinfo{person}{Krutika Parvatikar}, \bibinfo{person}{Marcos Zampieri}, \bibinfo{person}{Ashiqur~R. KhudaBukhsh}, {and} \bibinfo{person}{Liviu~P. Dinu}.} \bibinfo{year}{2025}\natexlab{}.
\newblock \showarticletitle{On the State of NLP Approaches to Modeling Depression in Social Media: A Post-COVID-19 Outlook}.
\newblock \bibinfo{journal}{\emph{IEEE Journal of Biomedical and Health Informatics}} \bibinfo{volume}{1}, \bibinfo{number}{1} (\bibinfo{year}{2025}), \bibinfo{pages}{1--13}.
\newblock
\href{https://doi.org/10.1109/JBHI.2025.3540507}{doi:\nolinkurl{10.1109/JBHI.2025.3540507}}


\bibitem[Cooper(2018)]%
        {cooper2018diagnosing}
\bibfield{author}{\bibinfo{person}{Rachel Cooper}.} \bibinfo{year}{2018}\natexlab{}.
\newblock \bibinfo{booktitle}{\emph{Diagnosing the Diagnostic and Statistical Manual of Mental Disorders}}.
\newblock \bibinfo{publisher}{Routledge}, \bibinfo{address}{London}.
\newblock
\showISBNx{9780429473678}
\href{https://doi.org/10.4324/9780429473678}{doi:\nolinkurl{10.4324/9780429473678}}


\bibitem[Gulliver et~al\mbox{.}(2010)]%
        {gulliver2010perceived}
\bibfield{author}{\bibinfo{person}{Amelia Gulliver}, \bibinfo{person}{Kathleen~M. Griffiths}, {and} \bibinfo{person}{Helen Christensen}.} \bibinfo{year}{2010}\natexlab{}.
\newblock \showarticletitle{Perceived barriers and facilitators to mental health help-seeking in young people: a systematic review}.
\newblock \bibinfo{journal}{\emph{BMC Psychiatry}} \bibinfo{volume}{10}, \bibinfo{number}{1} (\bibinfo{date}{30 Dec} \bibinfo{year}{2010}), \bibinfo{pages}{113}.
\newblock
\showISSN{1471-244X}
\href{https://doi.org/10.1186/1471-244X-10-113}{doi:\nolinkurl{10.1186/1471-244X-10-113}}


\bibitem[Guntuku et~al\mbox{.}(2017)]%
        {chandraguntuku2017detecting}
\bibfield{author}{\bibinfo{person}{Sharath~Chandra Guntuku}, \bibinfo{person}{David~B Yaden}, \bibinfo{person}{Margaret~L Kern}, \bibinfo{person}{Lyle~H Ungar}, {and} \bibinfo{person}{Johannes~C Eichstaedt}.} \bibinfo{year}{2017}\natexlab{}.
\newblock \showarticletitle{Detecting depression and mental illness on social media: an integrative review}.
\newblock \bibinfo{journal}{\emph{Current Opinion in Behavioral Sciences}}  \bibinfo{volume}{18} (\bibinfo{year}{2017}), \bibinfo{pages}{43--49}.
\newblock
\showISSN{2352-1546}
\href{https://doi.org/10.1016/j.cobeha.2017.07.005}{doi:\nolinkurl{10.1016/j.cobeha.2017.07.005}}
\newblock
\shownote{Big data in the behavioural sciences}.


\bibitem[Hauser et~al\mbox{.}(2022)]%
        {uhauser2022promise}
\bibfield{author}{\bibinfo{person}{Tobias~U Hauser}, \bibinfo{person}{Vasilisa Skvortsova}, \bibinfo{person}{Munmun {De Choudhury}}, {and} \bibinfo{person}{Nikolaos Koutsouleris}.} \bibinfo{year}{2022}\natexlab{}.
\newblock \showarticletitle{The promise of a model-based psychiatry: building computational models of mental ill health}.
\newblock \bibinfo{journal}{\emph{The Lancet Digital Health}} \bibinfo{volume}{4}, \bibinfo{number}{11} (\bibinfo{year}{2022}), \bibinfo{pages}{e816--e828}.
\newblock
\showISSN{2589-7500}
\href{https://doi.org/10.1016/S2589-7500(22)00152-2}{doi:\nolinkurl{10.1016/S2589-7500(22)00152-2}}


\bibitem[Husseini~Orabi et~al\mbox{.}(2018)]%
        {husseiniorabi2018deep}
\bibfield{author}{\bibinfo{person}{Ahmed Husseini~Orabi}, \bibinfo{person}{Prasadith Buddhitha}, \bibinfo{person}{Mahmoud Husseini~Orabi}, {and} \bibinfo{person}{Diana Inkpen}.} \bibinfo{year}{2018}\natexlab{}.
\newblock \showarticletitle{Deep Learning for Depression Detection of {T}witter Users}. In \bibinfo{booktitle}{\emph{Proceedings of the Fifth Workshop on Computational Linguistics and Clinical Psychology: From Keyboard to Clinic}}, \bibfield{editor}{\bibinfo{person}{Kate Loveys}, \bibinfo{person}{Kate Niederhoffer}, \bibinfo{person}{Emily Prud{'}hommeaux}, \bibinfo{person}{Rebecca Resnik}, {and} \bibinfo{person}{Philip Resnik}} (Eds.). \bibinfo{publisher}{Association for Computational Linguistics}, \bibinfo{address}{New Orleans, LA}, \bibinfo{pages}{88--97}.
\newblock
\href{https://doi.org/10.18653/v1/W18-0609}{doi:\nolinkurl{10.18653/v1/W18-0609}}


\bibitem[Lin(2004)]%
        {lin2004rouge}
\bibfield{author}{\bibinfo{person}{Chin-Yew Lin}.} \bibinfo{year}{2004}\natexlab{}.
\newblock \showarticletitle{{ROUGE}: A Package for Automatic Evaluation of Summaries}. In \bibinfo{booktitle}{\emph{Text Summarization Branches Out}}. \bibinfo{publisher}{Association for Computational Linguistics}, \bibinfo{address}{Barcelona, Spain}, \bibinfo{pages}{74--81}.
\newblock
\urldef\tempurl%
\url{https://aclanthology.org/W04-1013/}
\showURL{%
\tempurl}


\bibitem[Mohammad and Turney(2012)]%
        {mohammad2012crowdsourcing}
\bibfield{author}{\bibinfo{person}{Saif~M. Mohammad} {and} \bibinfo{person}{Peter~D. Turney}.} \bibinfo{year}{2012}\natexlab{}.
\newblock \showarticletitle{CROWDSOURCING A WORD–EMOTION ASSOCIATION LEXICON}.
\newblock \bibinfo{journal}{\emph{Computational Intelligence}} \bibinfo{volume}{29}, \bibinfo{number}{3} (\bibinfo{date}{Sept.} \bibinfo{year}{2012}), \bibinfo{pages}{436–465}.
\newblock
\showISSN{1467-8640}
\href{https://doi.org/10.1111/j.1467-8640.2012.00460.x}{doi:\nolinkurl{10.1111/j.1467-8640.2012.00460.x}}


\bibitem[Papineni et~al\mbox{.}(2002)]%
        {papineni2002bleu}
\bibfield{author}{\bibinfo{person}{Kishore Papineni}, \bibinfo{person}{Salim Roukos}, \bibinfo{person}{Todd Ward}, {and} \bibinfo{person}{Wei-Jing Zhu}.} \bibinfo{year}{2002}\natexlab{}.
\newblock \showarticletitle{BLEU: a method for automatic evaluation of machine translation}. In \bibinfo{booktitle}{\emph{Proceedings of the 40th Annual Meeting on Association for Computational Linguistics}} (Philadelphia, Pennsylvania) \emph{(\bibinfo{series}{ACL '02})}. \bibinfo{publisher}{Association for Computational Linguistics}, \bibinfo{address}{USA}, \bibinfo{pages}{311–318}.
\newblock
\href{https://doi.org/10.3115/1073083.1073135}{doi:\nolinkurl{10.3115/1073083.1073135}}


\bibitem[P{\'e}rez et~al\mbox{.}(2025)]%
        {prez2025depresym}
\bibfield{author}{\bibinfo{person}{Anxo P{\'e}rez}, \bibinfo{person}{Marcos Fern{\'a}ndez-Pichel}, \bibinfo{person}{Javier Parapar}, {and} \bibinfo{person}{David~E. Losada}.} \bibinfo{year}{2025}\natexlab{}.
\newblock \showarticletitle{DepreSym: A Depression Symptom Annotated Corpus and the Role of Large Language Models as Assessors of Psychological Markers}.
\newblock \bibinfo{journal}{\emph{Language Resources and Evaluation}} \bibinfo{volume}{59}, \bibinfo{number}{2} (\bibinfo{date}{03 May} \bibinfo{year}{2025}), \bibinfo{pages}{1--26}.
\newblock
\showISSN{1574-0218}
\href{https://doi.org/10.1007/s10579-025-09831-6}{doi:\nolinkurl{10.1007/s10579-025-09831-6}}


\bibitem[{Pew Research Center}(2024)]%
        {researchcenter2024how}
\bibfield{author}{\bibinfo{person}{{Pew Research Center}}.} \bibinfo{year}{2024}\natexlab{}.
\newblock \bibinfo{title}{How Americans Use Social Media}.
\newblock
\urldef\tempurl%
\url{https://www.pewresearch.org/internet/2024/01/31/americans-social-media-use/}
\showURL{%
\tempurl}


\bibitem[Picardi et~al\mbox{.}(2016)]%
        {picardi2016randomised}
\bibfield{author}{\bibinfo{person}{A. Picardi}, \bibinfo{person}{I. Lega}, \bibinfo{person}{L. Tarsitani}, \bibinfo{person}{M. Caredda}, \bibinfo{person}{G. Matteucci}, \bibinfo{person}{M.P. Zerella}, \bibinfo{person}{R. Miglio}, \bibinfo{person}{A. Gigantesco}, \bibinfo{person}{M. Cerbo}, \bibinfo{person}{A. Gaddini}, \bibinfo{person}{F. Spandonaro}, \bibinfo{person}{M. Biondi}, {and} \bibinfo{person}{{The SET-DEP Group}}.} \bibinfo{year}{2016}\natexlab{}.
\newblock \showarticletitle{A randomised controlled trial of the effectiveness of a program for early detection and treatment of depression in primary care}.
\newblock \bibinfo{journal}{\emph{Journal of Affective Disorders}}  \bibinfo{volume}{198} (\bibinfo{year}{2016}), \bibinfo{pages}{96--101}.
\newblock
\showISSN{0165-0327}
\href{https://doi.org/10.1016/j.jad.2016.03.025}{doi:\nolinkurl{10.1016/j.jad.2016.03.025}}


\bibitem[Plutchik and Kellerman(1980)]%
        {plutchik1980emotion}
\bibfield{author}{\bibinfo{person}{Robert Plutchik} {and} \bibinfo{person}{Henry Kellerman}.} \bibinfo{year}{1980}\natexlab{}.
\newblock \showarticletitle{EMOTION: Theory, Research, and Experience}.
\newblock In \bibinfo{booktitle}{\emph{Theories of Emotion}}, \bibfield{editor}{\bibinfo{person}{Robert Plutchik} {and} \bibinfo{person}{Henry Kellerman}} (Eds.). \bibinfo{publisher}{Academic Press}, \bibinfo{address}{London}, \bibinfo{pages}{ii}.
\newblock
\showISBNx{978-0-12-558701-3}
\href{https://doi.org/10.1016/B978-0-12-558701-3.50001-6}{doi:\nolinkurl{10.1016/B978-0-12-558701-3.50001-6}}


\bibitem[Reimers and Gurevych(2019)]%
        {reimers2019sentence-bert}
\bibfield{author}{\bibinfo{person}{Nils Reimers} {and} \bibinfo{person}{Iryna Gurevych}.} \bibinfo{year}{2019}\natexlab{}.
\newblock \showarticletitle{Sentence-{BERT}: Sentence Embeddings using {S}iamese {BERT}-Networks}. In \bibinfo{booktitle}{\emph{Proceedings of the 2019 Conference on Empirical Methods in Natural Language Processing and the 9th International Joint Conference on Natural Language Processing (EMNLP-IJCNLP)}}, \bibfield{editor}{\bibinfo{person}{Kentaro Inui}, \bibinfo{person}{Jing Jiang}, \bibinfo{person}{Vincent Ng}, {and} \bibinfo{person}{Xiaojun Wan}} (Eds.). \bibinfo{publisher}{Association for Computational Linguistics}, \bibinfo{address}{Hong Kong, China}, \bibinfo{pages}{3982--3992}.
\newblock
\href{https://doi.org/10.18653/v1/D19-1410}{doi:\nolinkurl{10.18653/v1/D19-1410}}


\bibitem[Wei et~al\mbox{.}(2022)]%
        {wei2022chain-of-thought}
\bibfield{author}{\bibinfo{person}{Jason Wei}, \bibinfo{person}{Xuezhi Wang}, \bibinfo{person}{Dale Schuurmans}, \bibinfo{person}{Maarten Bosma}, \bibinfo{person}{Brian Ichter}, \bibinfo{person}{Fei Xia}, \bibinfo{person}{Ed~H. Chi}, \bibinfo{person}{Quoc~V. Le}, {and} \bibinfo{person}{Denny Zhou}.} \bibinfo{year}{2022}\natexlab{}.
\newblock \showarticletitle{Chain-of-thought prompting elicits reasoning in large language models}. In \bibinfo{booktitle}{\emph{Proceedings of the 36th International Conference on Neural Information Processing Systems}} (New Orleans, LA, USA) \emph{(\bibinfo{series}{NIPS '22})}. \bibinfo{publisher}{Curran Associates Inc.}, \bibinfo{address}{Red Hook, NY, USA}, Article \bibinfo{articleno}{1800}, \bibinfo{numpages}{14}~pages.
\newblock
\showISBNx{9781713871088}


\bibitem[{World Health Organization}(2023)]%
        {healthorganization2023depressive}
\bibfield{author}{\bibinfo{person}{{World Health Organization}}.} \bibinfo{year}{2023}\natexlab{}.
\newblock \bibinfo{title}{Depressive disorder (depression)}.
\newblock
\urldef\tempurl%
\url{https://www.who.int/news-room/fact-sheets/detail/depression}
\showURL{%
\tempurl}
\newblock
\shownote{Fact sheet}.


\bibitem[Yang et~al\mbox{.}(2023)]%
        {yang2023towards}
\bibfield{author}{\bibinfo{person}{Kailai Yang}, \bibinfo{person}{Shaoxiong Ji}, \bibinfo{person}{Tianlin Zhang}, \bibinfo{person}{Qianqian Xie}, \bibinfo{person}{Ziyan Kuang}, {and} \bibinfo{person}{Sophia Ananiadou}.} \bibinfo{year}{2023}\natexlab{}.
\newblock \showarticletitle{Towards Interpretable Mental Health Analysis with Large Language Models}. In \bibinfo{booktitle}{\emph{Proceedings of the 2023 Conference on Empirical Methods in Natural Language Processing}}, \bibfield{editor}{\bibinfo{person}{Houda Bouamor}, \bibinfo{person}{Juan Pino}, {and} \bibinfo{person}{Kalika Bali}} (Eds.). \bibinfo{publisher}{Association for Computational Linguistics}, \bibinfo{address}{Singapore}, \bibinfo{pages}{6056--6077}.
\newblock
\href{https://doi.org/10.18653/v1/2023.emnlp-main.370}{doi:\nolinkurl{10.18653/v1/2023.emnlp-main.370}}


\bibitem[Zhang et~al\mbox{.}(2020)]%
        {zhang2020bertscore}
\bibfield{author}{\bibinfo{person}{Tianyi Zhang}, \bibinfo{person}{Varsha Kishore}, \bibinfo{person}{Felix Wu}, \bibinfo{person}{Kilian~Q. Weinberger}, {and} \bibinfo{person}{Yoav Artzi}.} \bibinfo{year}{2020}\natexlab{}.
\newblock \showarticletitle{BERTScore: Evaluating Text Generation with {BERT}}. In \bibinfo{booktitle}{\emph{8th International Conference on Learning Representations (ICLR)}}. \bibinfo{publisher}{OpenReview.net}, \bibinfo{address}{Virtual Conference}, \bibinfo{pages}{1--9}.
\newblock
\urldef\tempurl%
\url{https://openreview.net/forum?id=SkeHuCVFDr}
\showURL{%
\tempurl}
\newblock
\shownote{Conference held virtually}.


\bibitem[Zheng et~al\mbox{.}(2023)]%
        {zheng2023judging}
\bibfield{author}{\bibinfo{person}{Lianmin Zheng}, \bibinfo{person}{Wei-Lin Chiang}, \bibinfo{person}{Ying Sheng}, \bibinfo{person}{Siyuan Zhuang}, \bibinfo{person}{Zhanghao Wu}, \bibinfo{person}{Yonghao Zhuang}, \bibinfo{person}{Zi Lin}, \bibinfo{person}{Zhuohan Li}, \bibinfo{person}{Dacheng Li}, \bibinfo{person}{Eric~P. Xing}, \bibinfo{person}{Hao Zhang}, \bibinfo{person}{Joseph~E. Gonzalez}, {and} \bibinfo{person}{Ion Stoica}.} \bibinfo{year}{2023}\natexlab{}.
\newblock \showarticletitle{Judging LLM-as-a-judge with MT-bench and Chatbot Arena}. In \bibinfo{booktitle}{\emph{Proceedings of the 37th International Conference on Neural Information Processing Systems}} (New Orleans, LA, USA) \emph{(\bibinfo{series}{NIPS '23})}. \bibinfo{publisher}{Curran Associates Inc.}, \bibinfo{address}{Red Hook, NY, USA}, Article \bibinfo{articleno}{2020}, \bibinfo{numpages}{29}~pages.
\newblock


\end{thebibliography}

\end{document}